\documentclass[letterpaper]{article} 
\usepackage{aaai25}  
\usepackage{times}  
\usepackage{helvet}  
\usepackage{courier}  
\usepackage[hyphens]{url}  
\usepackage{graphicx} 
\usepackage{natbib}  
\usepackage{caption} 
\usepackage{algorithm}
\usepackage{algorithmic}
 
\usepackage{newfloat}
\usepackage{listings}
\DeclareCaptionStyle{ruled}{labelfont=normalfont,labelsep=colon,strut=off} 
\floatstyle{ruled}
\newfloat{listing}{tb}{lst}{}
\floatname{listing}{Listing}
\title{Twofold Debiasing Enhances Fine-Grained Learning with Coarse Labels}
\author{
Xin-yang Zhao\textsuperscript{\rm 1},
Jian Jin\textsuperscript{\rm 1},
Yang-yang Li\textsuperscript{\rm 1},
Yazhou Yao\textsuperscript{\rm 1}\thanks{Corresponding author.}
}

\affiliations{
    \textsuperscript{\rm 1}{School of Computer Science and
    Engineering, Nanjing University of Science and Technology,\\  
    Nanjing, Jiangsu, China}\\  
    
    \{zhaoxy, jinj, lyylyyi599, yazhou.yao\}@njust.edu.cn
}

\begin{document}

\maketitle

\begin{abstract}
The Coarse-to-Fine Few-Shot (C2FS) task is designed to train models using only coarse labels, then leverages a limited number of subclass samples to achieve fine-grained recognition capabilities. This task presents two main challenges: coarse-grained supervised pre-training suppresses the extraction of critical fine-grained features for subcategory discrimination, and models suffer from overfitting due to biased distributions caused by limited fine-grained samples. In this paper, we propose the Twofold Debiasing (TFB) method, which addresses these challenges through detailed feature enhancement and distribution calibration. Specifically, we introduce a multi-layer feature fusion reconstruction module and an intermediate layer feature alignment module to combat the model's tendency to focus on simple predictive features directly related to coarse-grained supervision, while neglecting complex fine-grained level details. Furthermore, we mitigate the biased distributions learned by the fine-grained classifier using readily available coarse-grained sample embeddings enriched with fine-grained information. Extensive experiments conducted on five benchmark datasets demonstrate the efficacy of our approach, achieving state-of-the-art results that surpass competitive methods.
\end{abstract}

%
\begin{links}
    \link{Code}{https://github.com/Faithzh/TFB}
\end{links}

\section{Introduction}

\begin{figure}[t]
    \centering
    \includegraphics[width=1\linewidth]{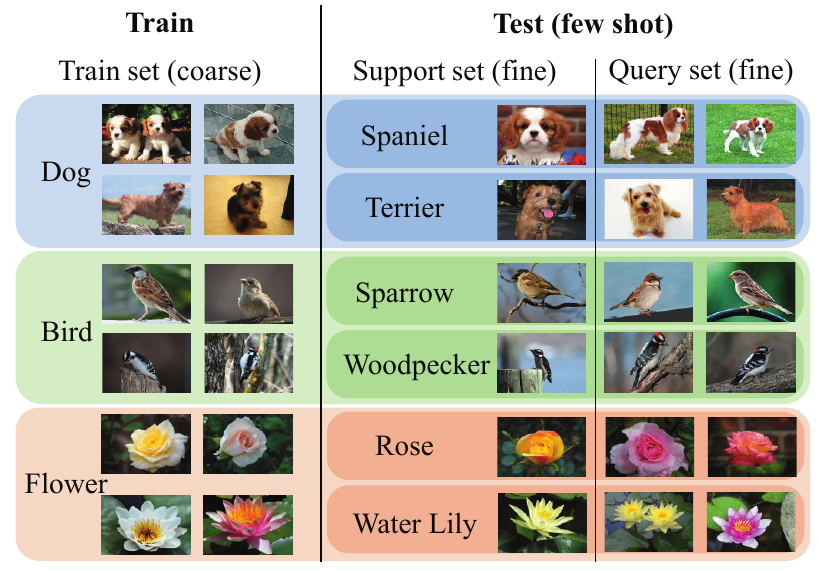}
    \caption{The Coarse-to-Fine Few-Shot (C2FS) task involves training a feature extractor using a training set with coarse labels, then the classifier leverages a support set with fine labels from a limited number of samples to achieve fine-grained recognition capabilities during testing.}
    \label{fig:task}
\end{figure}

Fine-grained recognition is a longstanding and fundamental problem in pattern recognition, underpinned by a diverse set of real-world applications~\cite{wei2021fine, shen2022semicon, xu2023hierarchical, shen2024equiangular, zhou2020bbn, chen2019multi, zhang2022guided, sun2021webly, yao2020bridging}. However, acquiring fine-grained labels in large quantities is challenging. On the one hand, this difficulty stems from the complexity of annotating with fine-grained knowledge; on the other hand, in many practical applications, the target label set of interest is not static and may change according to evolving needs. In contrast, coarse labels are more readily obtainable and often constitute the majority of dataset annotations. Therefore, Coarse-to-Fine Few-Shot (C2FS) task was proposed to exploit abundant coarse labels to enhance fine-grained recognition~\cite{bukchin2021fine}. The C2FS framework trains models on large-scale coarse-grained samples and then rapidly adapts them to finer recognition tasks using only a few fine-grained samples. As shown in Figure~\ref{fig:task}, unlike most previous few-shot tasks~\cite{ref26, wei2019piecewise}, the new classes in the C2FS task are fine-grained subclasses of pre-trained categories, contained within known coarse categories.

This task is challenging because coarse-grained labels usually represent broader, more generalized categories. Therefore, the model may learn features that lack intra-class differentiability in pre-training, making it ineffective at distinguishing subtle differences between fine-grained categories. This mismatch in feature representation makes it challenging for the classifier to achieve satisfactory classification performance, even when trained on fine-grained labels. The predominant approach in C2FS effectively utilizes coarse-grained supervision to distinguish different coarse categories, and then combines this with contrastive self-supervised methods to learn distinctive intra-class features~\cite{bukchin2021fine,yang2021towards,touvron2021grafit,feng2023maskcon,xu2023hyperbolic}. In this paper, we follow this framework and propose a targeted approach for the C2FS task, termed \textbf{T}wo\textbf{F}old de\textbf{B}iasing enhances fine-grained learning (TFB).

\begin{figure}[t]
    \centering
    \includegraphics[width=0.8\linewidth]{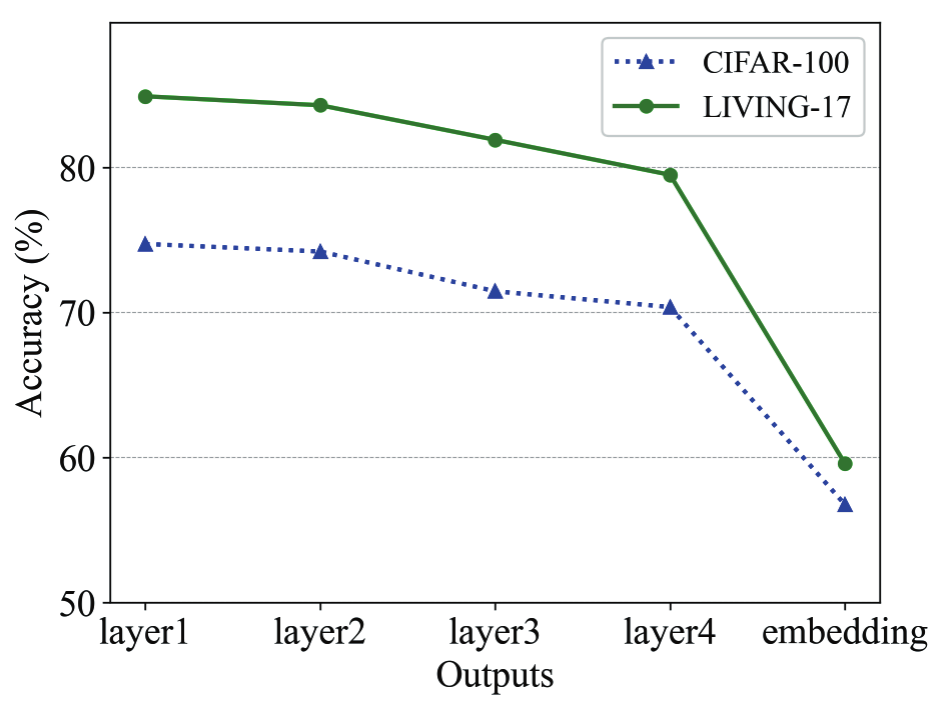}
    \caption{
    We froze the model trained under coarse-grained supervision and added a classification model to the output of each different layer. Then, these classification models were trained under fine-grained supervision to test their capability for fine-grained classification based on these different layer outputs.}
    \label{fig:cfacc}
\end{figure}
As shown in Figure~\ref{fig:cfacc}, we tested the fine-grained classification potential of outputs from different layers of a coarse-grained pre-trained model. The experiments reveal that under the influence of coarse-grained supervision, fine-grained semantic information gradually diminishes as the network deepens, with the most severe loss occurring in the deepest layer. The model tends to extract and emphasize semantic information directly related to the supervisory signals as the network depth increases while neglecting details that seem less directly relevant to the final task. 
In fact, although more fine-grained information could help distinguish between coarse categories, the model tends to focus on attributes that can most easily and directly differentiate between coarse categories. We refer to this phenomenon as the simplicity bias caused by coarse-grained supervision~\cite{simpBiasNIPs20}, i.e., neural networks simply ignore several complex predictive features in the presence of few simple predictive features. 

To mitigate the simplicity bias induced by coarse-grained supervision and improve the fine-grained feature recognition capabilities, we enhance the model's self-consistency. The model indirectly learns fine-grained feature representations by reconstructing the original image from the features outputted at the end, which contains all the detailed information~\cite{wei2023attribute}. Nevertheless, this approach might be suboptimal as the original image encompasses a vast amount of information irrelevant to fine-grained semantics, such as location, background, and pose. These redundant details could potentially be introduced into the model output. We address this by integrating features from different layers, compensating for the semantically irrelevant information during the reconstruction process. Based on our findings presented in Figure~\ref{fig:cfacc}, intermediate layer representations retain fine-grained discriminative features, even when directly influenced by coarse-grained supervisory constraints. More intuitively, the intermediate layer representations achieve a balance between preserving fine-grained discriminative attributes and excluding information irrelevant for coarse-grained classification. Building on this insight, we utilize these well-balanced intermediate representations to enhance the fine-grained representational capacity of the final embeddings.

More importantly, due to the limited availability of fine-grained samples, a good model finds it difficult to learn sufficient details from these sparse samples to accurately differentiate between highly similar categories. Notably, models pre-trained on coarse-grained supervision have already outperformed models pre-trained with fine-grained supervision. Especially on CIFAR-100, the previously best method~\cite{xu2023hyperbolic} even significantly surpasses the fine upper-bound baseline, exceeding it by 5.89\% and 4.93\% under 5-way 1-shot and all-way 1-shot conditions respectively. Intuitively, the feature extractor, either under fine-grained label supervision or with the aid of pretext tasks, has developed effective fine-grained discriminative abilities. But classifier trained with sparse data tend to learn a biased distribution, which does not fully reflect the true capabilities of the feature extractor. Previous works have focused on the feature extraction capabilities of backbones, neglecting the equally crucial role of the classifier. By exploiting the intrinsic relationships between base and target class representations, we aim to mitigate the distributional bias of target classes with limited samples by transferring statistical knowledge from richly-sampled base classes. By using the information from the base class samples, we enable the classifier to more accurately reveal the ground-truth data distribution, thereby learning a well-generalized model. 

The primary contributions of this work can be summarized as follows:
\begin{itemize}
\item[$\bullet$]We effectively enhance the fine-grained representational capability of features through multi-layer feature fusion reconstruction and intermediate layer feature alignment, thereby mitigating the simplicity bias in features under coarse-grained supervision.
\item[$\bullet$]Our analysis reveals that the distribution bias stemming from limited samples substantially constrains the model's discriminative capability. To mitigate this issue, we exploit the intrinsic relationships between base and target classes to rectify the distribution bias inherent in fine-grained few-shot scenarios.
\item[$\bullet$]Extensive experiments have demonstrated the effectiveness of TFB across multiple datasets, achieving state-of-the-art accuracy. Our approach offers a new solution for the C2FS task by simultaneously targeting the design of both the feature extractor and classifier.
\end{itemize}

\begin{figure*}[t]
    \centering
    \includegraphics[width=0.9\linewidth]{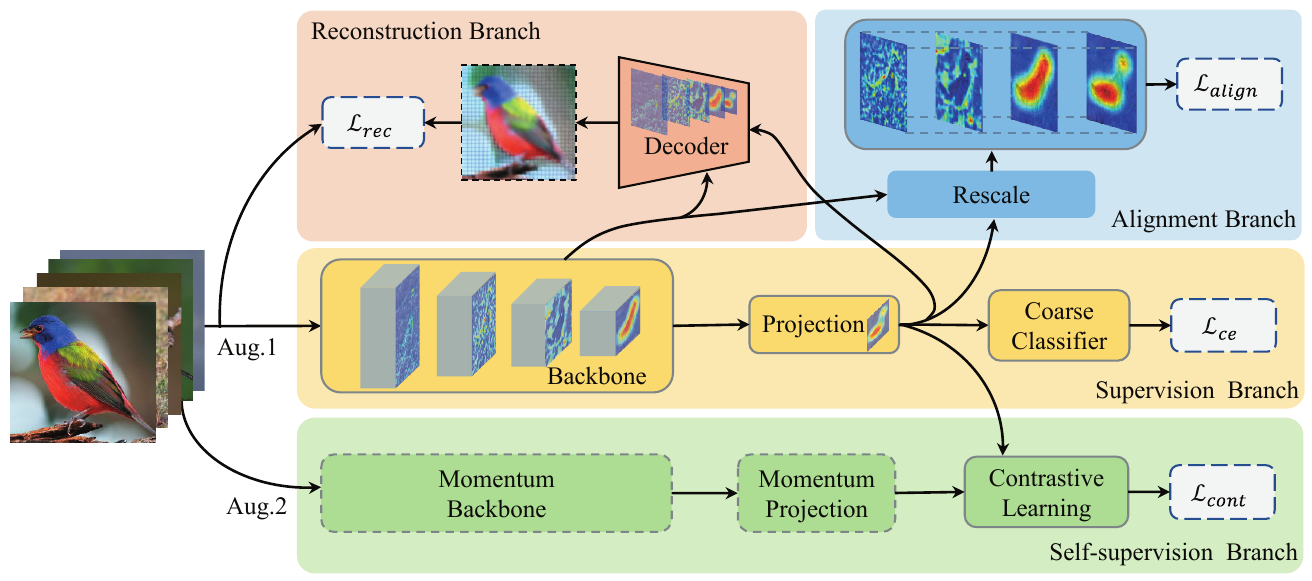}
    \caption{Overview of the TFB model training framework. The training objective consists of four components: coarse-grained cross-entropy loss $\mathcal{L}_{CE}$ for classifying coarse categories, reconstruction loss $\mathcal{L}_{rec}$ and alignment loss $\mathcal{L}_{align}$ to enhance the fine-grained representation capabilities of embeddings, and contrastive learning loss $\mathcal{L}_{cont}$ to learn better feature distributions within coarse category spaces. During testing, only the backbone is used as the feature extractor.}
    \label{fig:train}
    
\end{figure*}

\section{Related Work}
\label{sec:related work}

\subsection{Coarse-to-Fine Learning}

Coarse-to-Fine learning~\cite{ristin2015categories, taherkhani2019weakly, wei2023cat} has become a significant focus in computer vision and machine learning, aiming to leverage coarse-grained labeled data to enhance fine-grained recognition.
ANCOR~\cite{bukchin2021fine} introduced a fine-grained angular contrastive learning method that uses coarse labels to guide the angular loss function and generate sample pairs. Sun et al.~\cite{sun2021dynamic} proposed a dynamic metric learning approach that adapts the metric space to different semantic scales.
Yang~\cite{yang2021towards} address the cross-granularity gap by clustering coarse classes into pseudo-fine classes and introducing a meta-embedder that jointly optimizes both visual and semantic discrimination for effective pseudo-labeling.
Grafit~\cite{touvron2021grafit} implements a joint learning scheme that integrates instance and coarse label supervision losses to learn fine-grained image representations. 
MaskCon~\cite{feng2023maskcon} proposes a contrastive learning method that uses coarse labels to generate masked soft labels, leveraging both inter-sample relations and coarse label information. 
HCM~\cite{xu2023hyperbolic} embeds visual representations into a hyperbolic space and enhances their discriminative power using hierarchical cosine margins. 
FALCON~\cite{grcicfine} enables fine-grained class discovery from coarsely labeled data without requiring fine-grained level supervision.
Some studies~\cite{zhao2021mgsvf, xiang2022coarse} have also explored incremental few-shot learning with mixing of coarse and fine labels. 

\subsection{Self-supervised Learning}

As a branch of unsupervised learning, self-supervised learning (SSL) focuses on extracting discriminative features from large-scale unlabeled data, bypassing the need for human annotations~\cite{gui2024survey}. 
Context-based SSL methods exploit inherent contextual relationships, such as spatial structures and texture consistency within intact samples, using domain-specific pretext tasks~\cite{hu2024asymmetric} like image rotation prediction~\cite{gidaris2018unsupervised} and jigsaw puzzles~\cite{noroozi2016unsupervised}.
Contrastive learning has progressed from explicitly using negative examples~\cite{he2020momentum, chen2020improved} to ideas like self-distillation and feature decorrelation, all of which adhere to the principle of promoting positive example invariance while pushing apart different examples~\cite{wang2020understanding}. Masked image modeling~\cite{assran2022masked} reconstructs pixels~\cite{he2022masked} or local features~\cite{baevski2022data2vec}, leveraging co-occurrence relationships among image patches as supervision signals.

\subsection{Few-shot Learning}

Few-shot learning (FSL) is designed to mitigate the model's heavy reliance on large-scale training data, enabling models to quickly generalize to new tasks with limited annotated training data~\cite{10319790}. This recognition paradigm is ideally suited for fine-grained recognition, as collecting fine-grained samples is often costly. The methodologies for FSL can be taxonomically divided into three primary categories: data augmentation-based methods~\cite{ref26}, optimization-based methods~\cite{ref31}, and metric-based methods~\cite{jin2024few, wei2022embarrassingly}. 
Recent advances in metric-based methods have incorporated attention mechanisms~\cite{ref50} to model dependencies between elements in input sequences, utilized dense feature vectors to extract richer and finer image-to-image correlations~\cite{ref54}. While the aforementioned works have shown success in traditional FSL tasks, they still require a significant number of labels with the same granularity as the test data during training.

\section{Main Approach}

\subsection{Problem Definition}

In the study framework as detailed by ANCOR~\cite{bukchin2021fine}, define $\mathcal{Y}_{coarse}= \{y_1, \ldots ,y_N\}$ as a collection of $N$ broad training categories. We let ${\mathcal{S}_{train}^{coarse}=\{(I_i, y_i) | y_i \in \mathcal{Y}_{coarse} \}_{i=1}^M}$ represent a compilation of $M$ training images, each marked solely with labels from $\mathcal{Y}_{coarse}$. Consider $\mathcal{Y}_{fine}=\{y'_{1,1},\ldots,y'_{1,k_1}, \ldots ,y'_{N,1},\ldots, y'_{N,k_N}\}$ as a collection of fine-grained subclasses under the broad categories $\mathcal{Y}_{coarse}$. The encoder $E$ trained by $\mathcal{S}_{train}^{coarse}$ maps images $I$ into a $d$-dimensional feature space $\mathcal{F}$. At the testing phase, given a $k$-shot support set $\mathcal{S}_{sup}^{fine}=\{(I_r, y'_r) | y'_r \in \mathcal{Y}_{fine}^w\}_{r=1}^{k \cdot w}$ for a subset $\mathcal{Y}_{fine}^w \subseteq \mathcal{Y}_{fine}$ comprising $w$ fine categories, our objective is to train a classifier $ C_{fine} : \mathcal{F} \to \mathcal{Y}_{fine}^w$ to achieve optimal accuracy on the query set of $\mathcal{Y}_{fine}^w$ fine categories.

\subsection{Embedding Learning Debias}

The objective of this work is to learn an embedding function $f_{embed}(\cdot)$ that leverages coarsely-labeled training data to project input samples into a feature space with enhanced fine-grained discriminability. As shown in Figure~\ref{fig:train}, our architecture is built upon the self-supervised learning framework MoCo~\cite{he2020momentum}, consisting of a feature encoder $E$ with global average pooling, an projector $P$, a second pair of momentum encoder $E_m$ and momentum projector $P_m$ for encoding the positive keys in the contrastive objective that are momentum-updated from $E$ and $P$ and a coarse linear classifier $C_{coarse}$~\cite{bukchin2021fine}. Our proposed framework incorporates two essential components: a decoder $D$ for input reconstruction through multi-layer feature fusion, and an alignment module for bridging the intermediate representations and final embeddings. 

Firstly, we use coarse-grained labels $y$ for supervised learning, enabling the model to initially acquire coarse-grained recognition capabilities. Using $I_q$ to represent the input after data augmentation, we can obtain:
\begin{equation}\label{eq:loss_ce}
    \mathcal{L}_{CE}= CrossEntropy(C_{coarse}(P(E(I_q))), y)\,.
\end{equation}

\begin{figure}[t]
    \centering
    \includegraphics[width=0.95\linewidth]{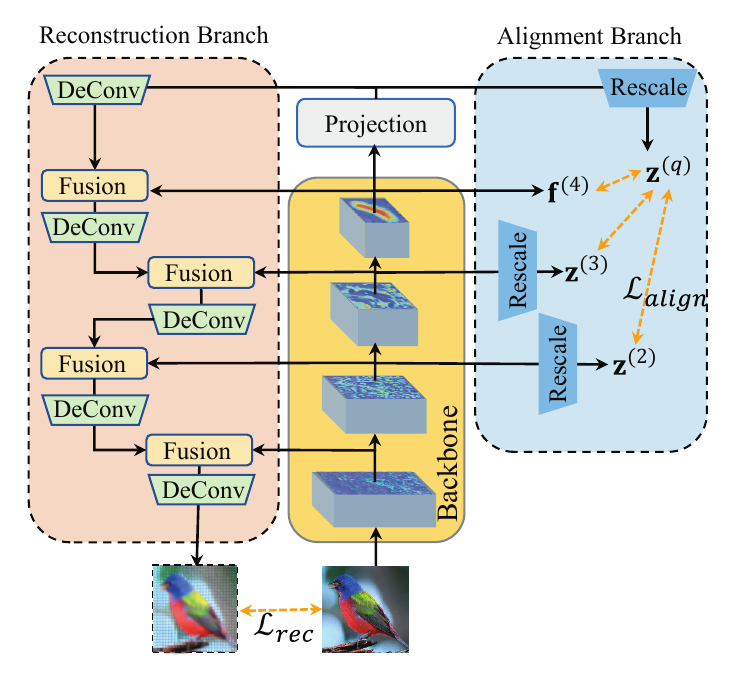}
    \caption{Multi-layer feature fusion for reconstruction branch and Intermediate layer feature alignment branch. The proposed fusion module employs cascaded concatenation operations and convolution layers to facilitate adaptive feature aggregation across different scales. $\mathbf{z}^{(i)}$ represents the rescaled $\mathbf{f}^{(i)}$.}
    \label{fig:rec}
\end{figure}

\begin{figure*}[t]
    \centering
    \includegraphics[width=0.9\linewidth]{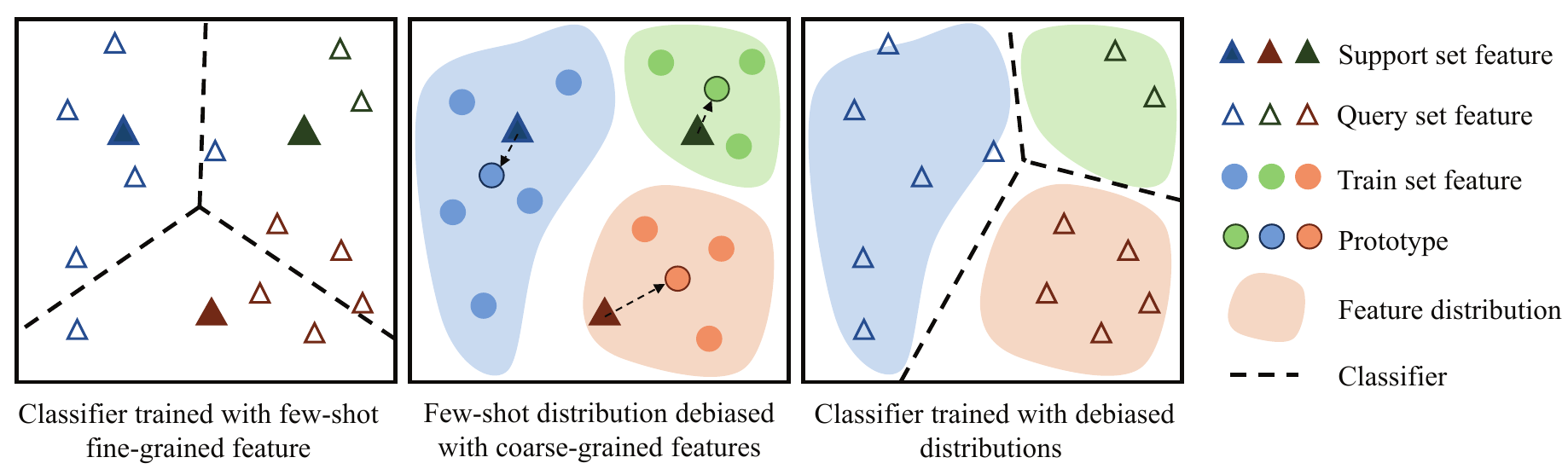}
    \caption{Training a classifier with limited fine-grained samples leads to overfitting (left). Correcting the distribution of few-shot samples in the feature space using training set features (middle). The classifier trained from the calibrated distribution features has better generalization capabilities (right).}
    \label{fig:test}
\end{figure*}

\subsubsection{Multi-layer Feature Fusion for Reconstruction}

Reconstruction branch aims to alleviate the simplicity bias caused by coarse-grained supervision by enhancing the model's self-consistency~\cite{ma2022principles}. Self-consistency compels the model to learn a variety of features ranging from basic textures to complex objects, thereby achieving a more comprehensive feature representation~\cite{wei2023attribute}. Specifically, the model indirectly learns fine-grained feature representations by reconstructing the original image $I_q$, which contains all the detail information. However, during the reconstruction process, it is detrimental for the model to focus on task-irrelevant information and structures within the input image $I_q$. Therefore, we reconstruct the image by merging features from different layers, utilizing shallow layer features to provide low-level features such as background and positional information, allowing the deeper layers to focus on semantic reconstruction.
More specifically, the encoder $E$ transforms an input image $I_q$ into a set of latent representations across multiple levels of abstraction in different layers. Let $\mathbf{f}^{(l)}$ represent the feature map at layer $l$ , where $l$ ranges from the shallow to the deep layers within the encoder, capturing varying degrees of semantic and structural details:
\begin{equation}
    \mathbf{f}^{(l)} = E^{(l)}(I_q) \,.
\end{equation}

Then, a deconvolutional network $D$~\cite{deconvCVPR} is applied upon $\mathbf{f}^{(l)}$ and $\mathbf{f}_q$, which is expected to reconstruct the input raw image $I_q$. As detailedly illustrated in Figure~\ref{fig:rec}, this reconstruction is achieved by integrating feature maps $ \mathbf{f}^{(l)} $ from multiple layers:
\begin{equation}
    \hat{I} = D(\mathbf{f}^{(1)}, \mathbf{f}^{(2)}, \ldots , \mathbf{f}^{q})\,,
\end{equation}
where $\mathbf{f}^{q}$ is the embedding of $I_q$. We used the MSE loss as the reconstruction loss $ \mathcal{L}_{rec} $ to guide the learning process in faithfully regenerating the original image. 

\subsubsection{Intermediate Layer Feature Alignment}

In the experiments shown in Figure~\ref{fig:cfacc}, we found that the information of fine-grained categories under coarse-grained supervision is mainly lost in the embeddings of the deepest layer outputs, while the model’s intermediate layer outputs still perform well in fine-grained recognition. As shown in Figure~\ref{fig:rec}, a feature alignment module is designed to utilize intermediate layer features for self-knowledge distillation. Each intermediate feature map $\mathbf{f}^{(i)}$ and the final embedding $\mathbf{f}^{q}$ are rescaled to match the spatial dimensions of the reference layer feature map $\mathbf{f}^{(4)}$. This rescaling is denoted by a function $Rescale(\mathbf{f}, s)$, where $s$ represents the target scale size derived from $\mathbf{f}^{(4)}$. The function $Rescale(\mathbf{f}, s)$ is actually composed of no more than two layers of convolution or deconvolution. The rescaled feature maps are then aligned through MSE loss minimization. The alignment process effectively acts as a form of regularization, encouraging the network to maintain a consistent representation of details across its depth. The overall align loss $ \mathcal{L}_{align} $ is
\begin{equation}
    \mathcal{L}_{align} = \sum_{i=2}^{4} \| Rescale(\mathbf{f}^{(i)}, s) - Rescale(\mathbf{f}^{q}, s) \|_2^2 \,.
\end{equation}

Although fine-grained information is enriched through image reconstruction and feature alignment, features may still exhibit arbitrary distributions within coarse category regions, potentially hindering discrimination between fine-grained categories. Therefore, we adopt the contrastive learning strategy $\mathcal{L}_{cont}$ proposed in ANCOR~\cite{bukchin2021fine} to enhance the self-organization and separability of fine-grained features.  

The final training objective is formulated as:
\begin{equation}\label{eq:loss_all}
    \mathcal{L}= \mathcal{L}_{CE} + \mathcal{L}_{cont} + \alpha \mathcal{L}_{rec} + \beta \mathcal{L}_{align}\,,
\end{equation}
where $\alpha$ and $\beta$ are tuning parameters that balance the importance of different losses against the primary task performance. To simplify the tuning process, we set $\alpha = \beta$.

\subsection{Fine-grained Classifier Debias}

\begin{table*}[t]
\begin{center}
\tabcolsep=0.1cm
\resizebox{1\linewidth}{!}{
\begin{tabular}{c|cccccccc}
\hline 
           {Methods}    & \multicolumn{2}{c}{LIVING-17}                     & \multicolumn{2}{c}{NONLIVING-26}                  & \multicolumn{2}{c}{ENTITY-13}   & \multicolumn{2}{c}{ENTITY-30}   \\ \cline{2-9} 
         & 5-way                   & all-way                 & 5-way                   & all-way                 & 5-way          & all-way        & 5-way          & all-way        \\ \hline
Fine upper-bound           & 90.75{\scriptsize$\pm$0.48}          & 62.65{\scriptsize$\pm$0.18}          & 90.33{\scriptsize$\pm$0.47}          & 60.68{\scriptsize$\pm$0.14}          & 94.72{\scriptsize$\pm$0.33} & 65.18{\scriptsize$\pm$0.09} & 94.02{\scriptsize$\pm$0.36} & 63.72{\scriptsize$\pm$0.10} \\
\hline
MoCo-v2          & 56.66{\scriptsize$\pm$0.70}          & 18.57{\scriptsize$\pm$0.11}          & 63.51{\scriptsize$\pm$0.75}          & 21.07{\scriptsize$\pm$0.11}          & 82.00{\scriptsize$\pm$0.67} & 33.06{\scriptsize$\pm$0.07} & 80.37{\scriptsize$\pm$0.62} & 28.62{\scriptsize$\pm$0.06} \\
MoCo-v2-ImageNet & 82.21{\scriptsize$\pm$0.73}          & 40.29{\scriptsize$\pm$0.14}          & 77.07{\scriptsize$\pm$0.78}          & 34.78{\scriptsize$\pm$0.13}          & 85.24{\scriptsize$\pm$0.60} & 35.62{\scriptsize$\pm$0.08} & 83.06{\scriptsize$\pm$0.62} & 31.73{\scriptsize$\pm$0.08} \\
SWAV-ImageNet   & 79.83{\scriptsize$\pm$0.65}          & 38.79{\scriptsize$\pm$0.15}          & 76.26{\scriptsize$\pm$0.71}          & 33.94{\scriptsize$\pm$0.11}          & 81.15{\scriptsize$\pm$0.65} & 33.57{\scriptsize$\pm$0.07} & 79.91{\scriptsize$\pm$0.54} & 31.15{\scriptsize$\pm$0.07} \\
ANCOR           & 89.23{\scriptsize$\pm$0.55}          & 45.14{\scriptsize$\pm$0.12}          & 86.23{\scriptsize$\pm$0.54}          & 43.10{\scriptsize$\pm$0.11}          & 90.58{\scriptsize$\pm$0.54} & 42.29{\scriptsize$\pm$0.08} & 88.12{\scriptsize$\pm$0.54} & 41.79{\scriptsize$\pm$0.08} \\
SCGM-G           & 89.72{\scriptsize$\pm$0.54}          & 48.74{\scriptsize$\pm$0.15}          & 89.87{\scriptsize$\pm$0.51}         & 49.25{\scriptsize$\pm$0.13}          & 90.15{\scriptsize$\pm$0.51} & 40.00{\scriptsize$\pm$0.08} & 92.90{\scriptsize$\pm$0.46} & 42.17{\scriptsize$\pm$0.08} \\ 
SCGM-A           & 90.97{\scriptsize$\pm$0.55}       & 49.31{\scriptsize$\pm$0.16}          & 88.78{\scriptsize$\pm$0.55}          & 46.93{\scriptsize$\pm$0.13}          & 88.48{\scriptsize$\pm$0.59} & 41.07{\scriptsize$\pm$0.09} & 91.22{\scriptsize$\pm$0.51} &  44.14{\scriptsize$\pm$0.09} \\
PE-HCM           &  90.94{\scriptsize$\pm$0.43} & 53.09{\scriptsize$\pm$0.11} & 89.97{\scriptsize$\pm$0.42} & \textbf{{50.12{\scriptsize$\pm$0.11}}} & 91.24{\scriptsize$\pm$0.38} & 41.64{\scriptsize$\pm$0.09} & \textbf{{92.95{\scriptsize$\pm$0.40}}} & 44.53{\scriptsize$\pm$0.09} \\
\hline
\textbf{Ours}            & \textbf{91.89{\scriptsize$\pm$0.54}} & \textbf{53.42{\scriptsize$\pm$0.12}} & \textbf{90.35{\scriptsize$\pm$0.53}} & 49.74{\scriptsize$\pm$0.11} & \textbf{91.56{\scriptsize$\pm$0.53}} & \textbf{43.23{\scriptsize$\pm$0.08}}  & 91.73{\scriptsize$\pm$0.52} & \textbf{45.03{\scriptsize$\pm$0.08}} \\ \hline
\end{tabular}
}
\end{center}
\caption{Accuracy comparisons of different methods on the BREEDS dataset under 5-way 1-shot and all-way 1-shot settings. Bold numbers are the best results.}
\label{tab:breeds}
\end{table*}

When trained on sparsely available fine-grained data, well-pretrained models also suffers from distribution bias and overfitting issues. Considering this, we aim to utilize the extensive information from coarse-grained training datasets to address these issues. If we can identify the relationships between training set samples and support set samples, we can calibrate the distribution of fine-grained classes using training set samples. Calibrating the distribution in the original data space is challenging, so we attempt to calibrate the distribution in the feature space. The feature space has far fewer dimensions than the original data space and is also easier to calibrate~\cite{xian2018feature,yang2021bridging}. Therefore, our approach is based on the premise that two semantically similar samples, once processed through a feature extractor, will yield proximate features in the feature space.

Following the feature extractor training, we establish a feature repository $\mathcal{T}$ by extracting and storing feature representations from the entire training set. Using $\mathcal{S}_{sup}^{y'}$ to represent the support set features corresponding to the fine-grained class $y'$, the prototype $ P^{y'} $~\cite{wei2022prototype} is initialized as the centroid of $ \mathcal{S}_{sup}^{y'}$. We first determine the coarse category $y$ to which $y'$ belongs. The $y$ is determined through a k-Nearest Neighbors (kNN) classification scheme operating on the training set feature repository $\mathcal{T}$. Specifically, for the prototype $ P^{y'} $, its coarse label $y$ is assigned based on majority voting among its $k$ nearest neighbors:
\begin{equation}
    y = {\mathop{\arg \max}}_{c} \sum_{i=1}^{k} \mathbf{1}(y_i = c) \,,
\end{equation}
where $y_i$ denotes the label of the $i$-th nearest neighbor in feature space, and $\mathbf{1}(\cdot)$ is the indicator function. Upon identifying the coarse label $ y $, a feature subset $ \mathcal{T}_y $ is extracted from $ \mathcal{T} $, consisting of embeddings labeled as $ y $. This subset provides a potential addition feature pool for $S_{sup}^{y'}$:
\begin{equation}
    \mathcal{T}_y = \{ x \in \mathcal{T} : label(x) = y \} \,.
\end{equation}

Then, an additional support set $\mathcal{S}_{add}^{y'}$ is constructed by selecting the $m$ nearest neighbors in feature space from $\mathcal{T}y$, where the distance is computed with $P^{y'}$.  Subsequently, the average features of $ S_{add}^{y'} $ are used to correct the original prototype $P^{y'}$:
\begin{equation}
    P^{y'} = \frac{1}{|\mathcal{S}_{sup}^{y'}|} \sum_{\mathbf{x} \in \mathcal{S}_{sup}^{y'}} \mathbf{x} + \frac{1}{|\mathcal{S}_{add}^{y'}|} \sum_{\mathbf{x} \in \mathcal{S}_{add}^{y'}} \mathbf{x} \,.
\end{equation}
This prototype calibration process needs to be repeated multiple times. The corrected prototype is then used to continue finding new $ m $ features to add to $ \mathcal{S}_{add}^{y'}$ from $ \mathcal{T}_y $ until the feature count in $\mathcal{S}_{add}^{y'}$ meets the requirement of $ n $ items.
Finally, the additional support set $\mathcal{S}_{add}^{y'}$ along with the original support set $\mathcal{S}_{sup}^{y'}$ are then served as the training data for fine-grained classifier.

\section{Experiments}

\subsection{Benchmark Datasets}
Our experiments were performed on BREEDS~\cite{taori2020breeds} and CIFAR-100~\cite{Krizhevsky2009} datasets. BREEDS is a collection of four datasets derived from ImageNet and recalibrated to ensure that classes at the same hierarchy level exhibit similar visual granularity. The CIFAR-100 dataset includes 20 coarse-grained classes and a total of 100 fine-grained classes.

\subsection{Baselines}
We compare our method with the most relevant state-of-the-art models on Coarse-to-Fine Few-Shot task: (1) MoCo-v2~\cite{chen2020improved}, trained on respective training sets. (2) MoCo-v2-ImageNet~\cite{chen2020improved}, represents the complete ImageNet pre-trained version of the model. (3) SWAV-ImageNet, the official model described in~\cite{caron2020swav}. (4) ANCOR~\cite{bukchin2021fine}, integrates supervised inter-class and self-supervised intra-class learning with an innovative angular normalization component. (5) SCGM~\cite{ni2022superclassconditional}, simulates the sample generation process using hierarchical class structures and explicitly incorporates latent variables to account for unobserved subclasses. (6) PE-HCM~\cite{xu2023hyperbolic}, embeds visual representations into hyperbolic space, enhancing their discriminative capacity through hierarchical cosine margins. (7) Fine upper-bound~\cite{bukchin2021fine}, naturally trained on fine-grained labels.

\subsection{Additional Implementation Details}
For fair comparisons with other methods, we use ResNet-12~\cite{he2016deep} as the backbone network for CIFAR-100~\cite{Krizhevsky2009} and ResNet-50~\cite{he2016deep} for BREEDS~\cite{taori2020breeds}. The output dimension of these backbone networks is $d = 2048$ or $d = 640$. In all experiments, the projector is implemented as a three-layer MLP with an output dimension of 128. The model is trained using the SGD optimizer on 4 GeForce RTX 3090 GPUs for 200 epochs. For CIFAR-100 and BREEDS, the batch sizes are 512 and 256 respectively; the initial learning rates are 0.12 and 0.03; the hyperparameters $\alpha$ are set at 1 and 10 respectively. The learning rates decrease by tenfold at the 140th and 180th epochs. We implement random data augmentation techniques including random resized crop, random horizontal flipping, and random color jitter during training. Other hyperparameter settings remain the same as those used in ANCOR. During the test phase, we evaluate the performance using 5-way and all-way 1-shot settings. The hyperparameters $k$, $m$ and $n$ are set to 10, 20 and 100 respectively. The evaluation is conducted on 1000 random episodes, and we report the mean accuracy along with a 95\% confidence interval.

\begin{table}[t]
\small
\begin{center}
\tabcolsep=0.15cm
\resizebox{0.75\linewidth}{!}{
\begin{tabular}{c|cc}
\hline
  Methods  & 5-way          & all-way        \\
\hline
Fine upper-bound   & 75.53{\scriptsize$\pm$0.68} & 31.35{\scriptsize$\pm$0.11} \\
\hline
ANCOR  & 74.56{\scriptsize$\pm$0.70} & 29.84{\scriptsize$\pm$0.11} \\
SCGM-G  & 76.19{\scriptsize$\pm$0.73} & 29.92{\scriptsize$\pm$0.11} \\
SCGM-A  & 77.37{\scriptsize$\pm$0.77} & 25.91{\scriptsize$\pm$0.10} \\
PE-HCM    & 81.42{\scriptsize$\pm$0.69} & 36.28{\scriptsize$\pm$0.12} \\
\hline
\textbf{Ours}    & \textbf{{84.15{\scriptsize$\pm$0.69}}} & \textbf{{39.03{\scriptsize$\pm$0.11}}} \\
\hline
\end{tabular}}
\end{center}
\caption{Comparisons on CIFAR-100~\cite{Krizhevsky2009}.  Bold numbers are the best results. }
\label{table:CIFAR}
\end{table}

\subsection{Main Result}

Following common experimental setups~\cite{Tian2020}, we present results for both 5-way 1-shot and all-way 1-shot configurations, with 15 queries per test episode. Table~\ref{tab:breeds} and table~\ref{table:CIFAR} report the average accuracy for the core use-case of the task evaluation. It is evident that our approach delivers superior performance across all datasets. Especially on the CIFAR-100 dataset, we observed an approximate 3\% significant advantage in accuracy. Traditional methods have focused on designing fine-grained pretext tasks to enable feature extractors to learn fine-grained feature distributions from coarse-grained datasets, subsequently verifying these distributions through fine-grained classifier. 

Our method introduces an alternative novel solution that not only preserves the fine-grained modeling capabilities of the feature extractor but also guides the fine-grained classifier to learn the correct distributions as much as possible. Compared to pretext-task-based approaches, our method offers two advantages: (1) Although only a limited amount of fine-grained data is used as the support set, this set provides accurate and authentic fine-grained category information. For example, as demonstrated by the PE-HCM method~\cite{xu2023hyperbolic}, the number of clusters significantly influences the performance of fine-grained recognition. (2) Features learned under coarse-grained labels calibrate the distribution bias in fine-grained classifiers for few-shot scenarios, effectively leveraging the relationship between the sparse fine-grained data in the support set and the abundant coarse-grained data in the training set, thereby bridging the gap between few-shot and many-shot conditions. 

\begin{table}[t]
\begin{center}
\tabcolsep=0.15cm
\resizebox{1\linewidth}{!}{
\begin{tabular}{ccc|cccc}
\hline
\multicolumn{3}{c|}{Methods}                                                      & \multicolumn{2}{c}{LIVING-17} & \multicolumn{2}{c}{CIFAR-100} \\ \hline
 REC                & ALIGN           & CD            & 5-way        & all-way        & 5-way        & all-way  \\
\hline
&   &  & 89.45    & 45.59     & 77.67   & 33.83          \\
 $\surd$ & & &90.74     & 47.32      & 80.05   & 36.80        \\
$\surd$ & $\surd$ & & 91.06      & 48.73      &   81.79   &     37.41   \\ 
$\surd$ & $\surd$ & $\surd$ & 91.89     & 53.42     & 84.15    & 39.03     \\ \hline
\end{tabular}}
\end{center}
\caption{Comparisons of our proposals on LIVING-17 and CIFAR-100. REC: Multi-layer feature fusion for reconstruction. ALIGN: Intermediate layer feature alignment. CD :Fine-grained classifier debias.}
\label{table:ablation}
\vspace{-1em}
\end{table}

\begin{table}[t]
\begin{center}
\tabcolsep=0.15cm
\resizebox{0.95\linewidth}{!}{
\begin{tabular}{c|cccc}
\hline
{Methods}     & \multicolumn{2}{c}{LIVING-17}                     & \multicolumn{2}{c}{CIFAR-100}        \\ \cline{2-5} 
         & 5-way                   & all-way                 & 5-way                   & all-way                       \\ \hline
All features& 91.02    & 51.71     & 83.19   & 37.61          \\
kNN-based& 91.89     & 53.42     & 84.15   & 39.03        \\
Coarse labeled& 92.91      & 54.95      &   86.92   &  43.22 \\ \hline
\end{tabular}}
\end{center}
\caption{Comparative experiment of different coarse-grained supervisions in the debiasing process of fine-grained classifiers. All features: completely disregards coarse granularity, obtaining  $\mathcal{S}_{add}^{y'}$  from all training features $\mathcal{T}$. kNN-based: uses a kNN classifier to obtain coarse categories in our method. Coarse labeled: the support set is annotated with coarse-grained labels.}
\label{table:coarse}
\vspace{-1em}
\end{table}

\subsection{Ablation Study}

The ablation analysis presented in table \ref{table:ablation} quantitatively demonstrates the contribution of each model component to the overall performance across two benchmark datasets. The results demonstrate that incorporating the REC component significantly improves accuracy, while the subsequent integration of the ALIGN component brings further enhancements. These two methods effectively maintain the fine-grained information preservation capability of the feature extractor. As there is no explicit modeling of fine-grained classification tasks, our model does not show a significant advantage over other methods before the use of CD. However, the CD strategy noticeably enhances the overall performance. This debiasing of the fine-grained classifier is crucial. In fact, our strategy has the potential to be applicable to previous methods. Overall, experimental results demonstrate that designing both the feature extractor and classifier specifically for the C2FS is effective and necessary.

\begin{figure}[t]
    \centering 
    \includegraphics[width=1\linewidth]{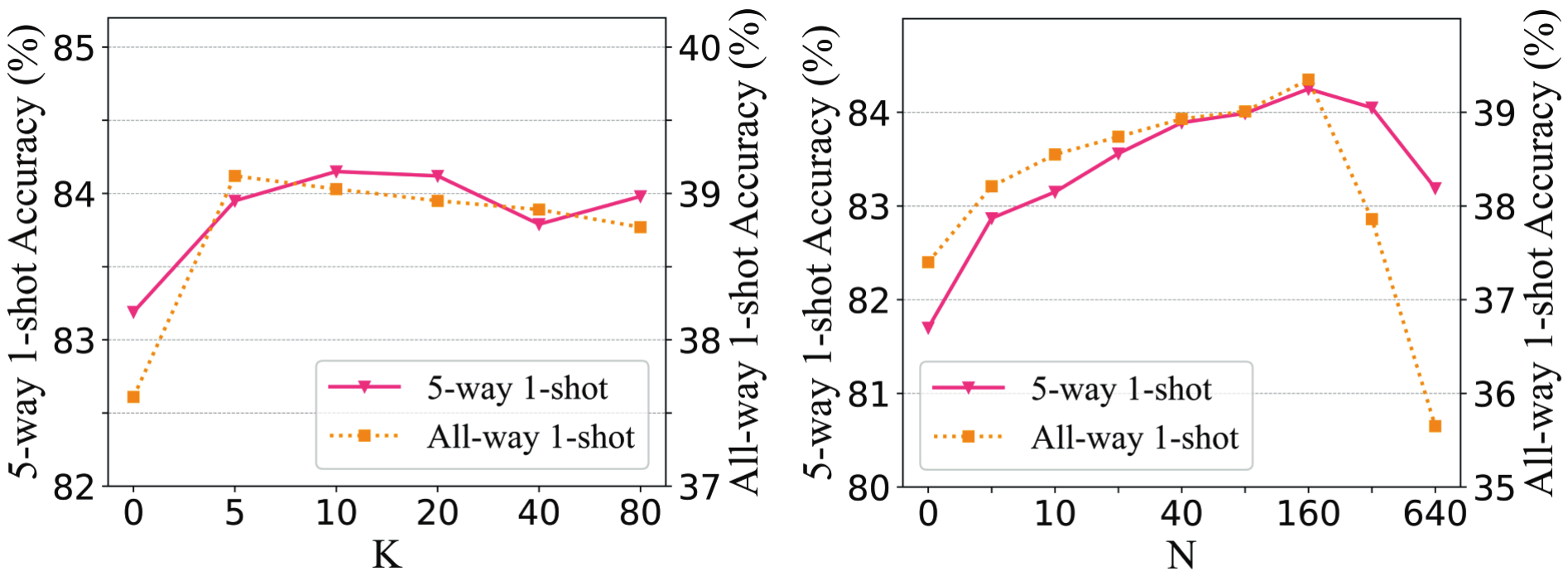}
    \caption{Ablation study of selected hyperparameters on the CIFAR-100 dataset. The Impact of Different kNN Parameters on Results(left). Influence of the Size $n$ of the Additional Support Set $\mathcal{S}_{add}^{y'}$ on Results(right).}
\label{fig:perm}
\end{figure}

Table~\ref{table:coarse} shows the effectiveness of incorporating coarse-grained labels in the debiasing process of fine-grained classifiers. The results demonstrate that coarse-grained label of the support set plays a crucial role in optimizing classifier performance. With coarse-grained labels available, the model more accurately identifies the corresponding feature subset $ \mathcal{T}_y $ for each category, resulting in substantial performance gains, notably a 6\% improvement in all-way accuracy on LIVING-17. When coarse-grained categories are disregarded, the CD strategy shows limited improvement. Evidently, utilizing the training set to additionally train a better coarse-grained classifier, instead of relying on simple kNN, can effectively guide the debiasing process.

Additionally, as illustrated in the left side of the Figure~\ref{fig:perm}, we conducted an ablation study on the kNN parameter $k$, finding that our method is not highly sensitive to $k$. On the right side of the Figure~\ref{fig:perm}, we experimented with different numbers of coarse granularity features $n$ in $\mathcal{S}_{add}^{y'}$. The results indicate that when $n$ is within a reasonable range, it can significantly improve model performance. However, when $n$ is sufficiently large, $\mathcal{S}_{add}^{y'}$ introduces a large number of false positive samples, leading to incorrect adjustments in the distribution.

\section{Conclusion}

In this paper, we introduce a novel methodology called TwoFold deBiasing (TFB) , which enhances fine-Grained learning for the C2FS task. Our approach mitigates the simplicity bias of feature extractors under coarse-grained supervision and tackles the biased distributions learned by fine-grained classifiers in few-shot scenarios. Specifically, TFB employs multi-layer feature fusion and intermediate layer feature alignment to enrich the detailed information required for fine-grained classification. Furthermore, by leveraging the hierarchical relationship between coarse and fine labels, TFB incorporates training set features to calibrate biased distributions in few-shot learning, enabling classifiers to better approximate the true data distribution. Our comprehensive experiments across benchmark datasets show the benefits of designing integrated training and testing strategies. 

\section{Acknowledgments}
This work was supported by the National Natural Science Foundation of China (No. 62472222), and Natural Science Foundation of Jiangsu Province (No. BK20240080).

\bibliography{aaai25}

\end{document}